\documentclass[letterpaper]{article}

\usepackage[noblocks]{authblk}
\usepackage{uai2018}
\usepackage{graphicx}
\usepackage{multirow}
\usepackage[margin=1in]{geometry}

\usepackage{times}

\usepackage{color,xcolor}
\usepackage{epsfig}
\usepackage{graphicx}

\usepackage{adjustbox}
\usepackage{array}
\usepackage{booktabs}
\usepackage{colortbl}
\usepackage{float,wrapfig}
\usepackage{hhline}
\usepackage{multirow}
\usepackage{subcaption} 

\usepackage{amsmath,amsfonts,amssymb}
\usepackage{bm}
\usepackage{nicefrac}
\usepackage{microtype}

\usepackage{changepage}
\usepackage{extramarks}
\usepackage{fancyhdr}
\usepackage{lastpage}
\usepackage{setspace}
\usepackage{soul}
\usepackage{xspace}

\usepackage{url}

\usepackage{algorithm, algorithmic}
\usepackage{enumerate}
\usepackage{todonotes} 
\usepackage{gensymb}

\usepackage{titlesec}

\newcolumntype{L}[1]{>{\raggedright\let\newline\\\arraybackslash\hspace{0pt}}m{#1}}
\newcolumntype{C}[1]{>{\centering\let\newline\\\arraybackslash\hspace{0pt}}m{#1}}
\newcolumntype{R}[1]{>{\raggedleft\let\newline\\\arraybackslash\hspace{0pt}}m{#1}}

\newcommand{\ignorethis}[1]{}

\makeatletter
\DeclareRobustCommand\onedot{\futurelet\@let@token\@onedot}
\def\@onedot{\ifx\@let@token.\else.\null\fi\xspace}

\def\eg{\emph{e.g}\onedot}

\def\etal{\emph{et al}\onedot}
\makeatother

\definecolor{MyDarkBlue}{rgb}{0,0.08,1}
\definecolor{MyDarkGreen}{rgb}{0.02,0.6,0.02}
\definecolor{MyDarkRed}{rgb}{0.8,0.02,0.02}
\definecolor{MyDarkOrange}{rgb}{0.40,0.2,0.02}
\definecolor{MyPurple}{RGB}{111,0,255}
\definecolor{MyRed}{rgb}{1.0,0.0,0.0}
\definecolor{MyGold}{rgb}{0.75,0.6,0.12}
\definecolor{MyDarkgray}{rgb}{0.66, 0.66, 0.66}

\newcommand{\myitem}{\vspace{-5pt}\item}

\titlespacing\section{2pt}{2pt plus 0pt minus 2pt}{2pt plus 0pt minus 2pt}
\titlespacing\subsection{1pt}{1pt plus 0pt minus 1pt}{1pt plus 0pt minus 1pt}

\title{Unsupervised Learning of Latent Physical Properties Using Perception-Prediction Networks}

\author[1]{David Zheng}
\author[2]{Vinson Luo}
\author[1]{Jiajun Wu}
\author[1]{Joshua B. Tenenbaum}
\affil[1]{Computer Science and Artificial Intelligence Laboratory, MIT}
\affil[2]{Department of Computer Science, Stanford University}

\begin{document}

\maketitle

\begin{abstract}

We propose a framework for the completely unsupervised learning of latent object properties from their interactions: the \textit{perception-prediction network} (PPN). Consisting of a perception module that extracts representations of latent object properties and a prediction module that uses those extracted properties to simulate system dynamics, the PPN can be trained in an end-to-end fashion purely from samples of object dynamics. The representations of latent object properties learned by PPNs not only are sufficient to accurately simulate the dynamics of systems comprised of previously unseen objects, but also can be translated directly into human-interpretable properties (\eg mass, coefficient of restitution) in an entirely unsupervised manner. Crucially, PPNs also generalize to 
novel scenarios: their gradient-based training can be applied to many dynamical systems and their graph-based structure functions over systems comprised of different numbers of objects. Our results demonstrate the efficacy of graph-based neural architectures in object-centric inference and prediction tasks, and our model has the potential to discover relevant object properties in systems that are not yet well understood.

\end{abstract}

\begin{figure*}[t]
    \centering
    \includegraphics[width=\textwidth]{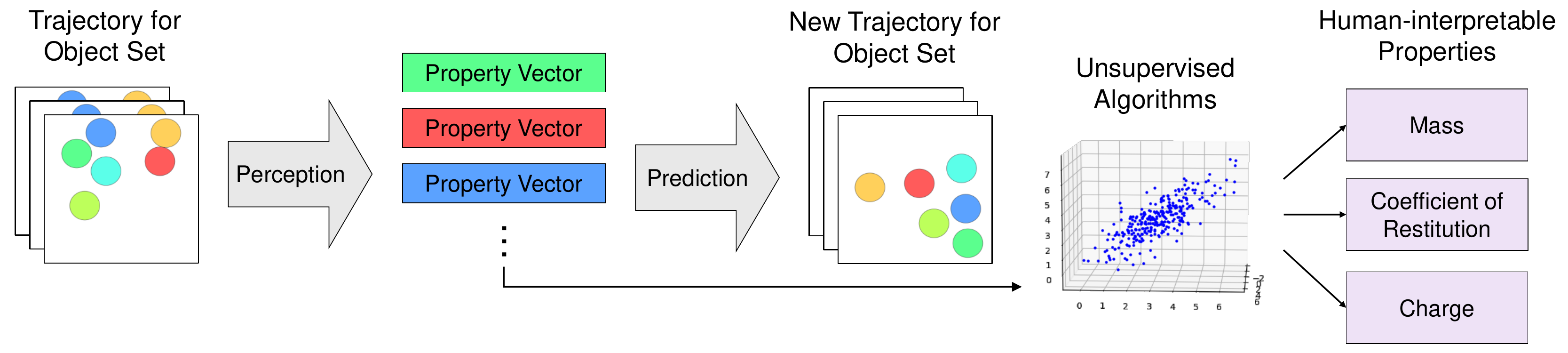}
    \caption{\textbf{Model overview.} The unsupervised object property discovery paradigm that the PPN follows extracts property vectors from samples of object dynamics to accurately predict new trajectories of those same objects. Applying unsupervised learning methods to the learned vectors allows for the extraction of human-interpretable object properties.}
    \vspace{-15pt}
    \label{fig:model_teaser}
\end{figure*}

\section{INTRODUCTION}

The physical properties of objects, combined with the laws of physics, govern the way in which objects move and interact in our world. Assigning properties to objects we observe helps us summarize our understanding of those objects and make better predictions of their future behavior. Often, the discovery of such properties can be performed with little supervision. For instance, by watching an archer shoot several arrows, we may conclude that properties such as the tension of the bowstring, the strength and direction of the wind, and the mass and drag coefficient of the arrow affect the arrow's ultimate trajectory. Even when given observations from entirely novel microworlds, humans are still able to learn the relevant physical properties that characterize a system \cite{ullman2014learning}.

Our work utilizes recent advances in neural relation networks in order to learn latent physical properties of a system in an unsupervised manner. In particular, the neural relation architectures~\cite{battaglia2016interaction,chang2016compositional} have proven capable of accurately simulating complex physical interactions involving objects with known physical properties. Relation networks have several characteristics that make them particularly suitable for our task: they are fully differentiable, allowing them to be applied to a variety of different situations without the need for any architectural change; they have a modular graph-based structure that generalizes over differing numbers of objects; and their basic architecture can be easily applied to both dynamics prediction and the learning of latent properties.

We use relation networks to construct the perception-prediction network (PPN), a novel system that uses a representation learning \cite{bengio2013representation} paradigm to extract an encoding of the properties of a physical system purely through observation. Unlike previous neural relation architectures, which only use relation networks to predict object states with known property values, we use relation networks to create both a \textit{perception network}, which derives property values from observations, and a \textit{prediction network}, which predicts object positions given property values. The PPN is able to derive unsupervised representations of the latent properties relevant to physical simulations purely by observing the dynamics of systems comprised of objects with different property values. These learned representations can be translated directly into human-interpretable properties such as mass and coefficient of restitution.

One crucial aspect of our system is generalization, which humans excel at when inferring latent properties of novel systems. Our proposed system is robust under several forms of generalization, and we present experiments demonstrating the ability of our unsupervised approach to discern interpretable properties even when faced with different numbers of objects during training and testing as well as property values in previously unseen ranges.

We evaluate the PPN for two major functionalities: the accuracy of dynamics prediction for unseen objects and the interpretability of properties learned by the model. We show that our model is capable of accurately simulating the dynamics of complex multi-interaction systems with unknown property values after only a short observational period to infer those property values. Furthermore, we demonstrate that the representations learned by our model can be easily translated into relevant human-interpretable properties using entirely unsupervised methods. Additionally, we use several experiments to show that both the accuracy of dynamics prediction and interpretability of properties generalize well to new scenarios with different numbers and configurations of objects. Ultimately, the PPN serves as a powerful and general framework for discovering underlying properties of a physical system and simulating its dynamics. 
\begin{figure*}[t]
    \centering
    \includegraphics[width=\textwidth]{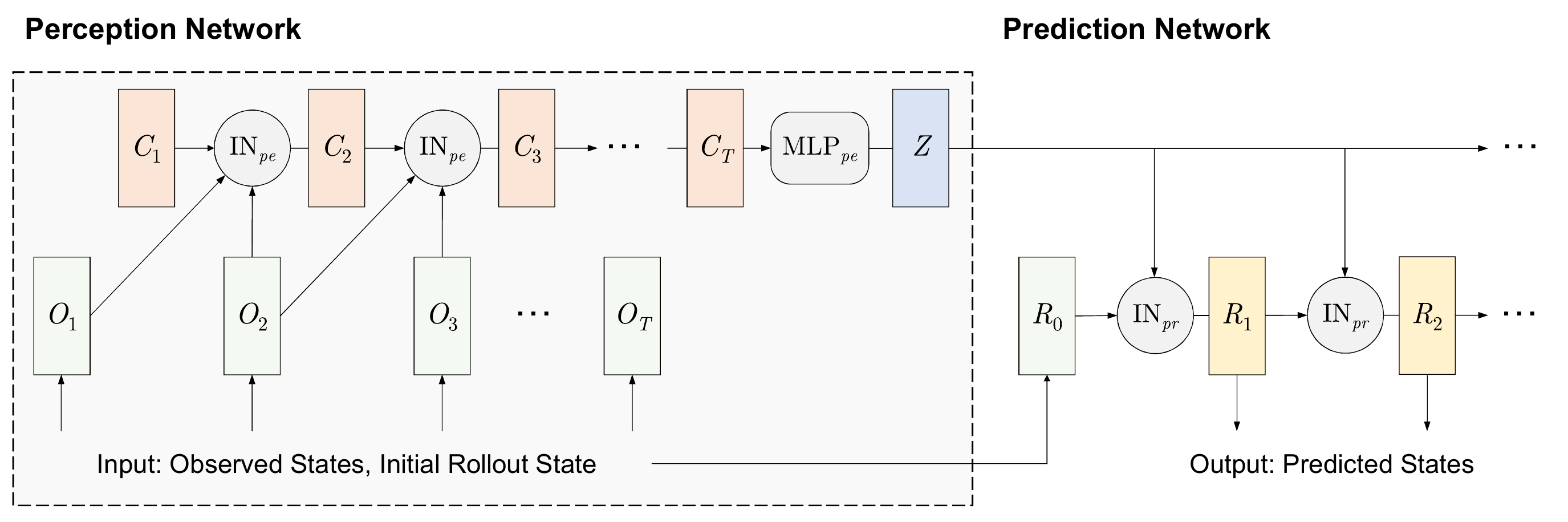}
    \caption{\textbf{Model architecture.} The PPN takes as input a sequence of observed states $O_1, \ldots, O_T$ as well an initial state $R_0$ to begin a new rollout. Code vectors $C_1, \ldots, C_T$ are derived from the observed states using interaction networks and a final property vector $Z$ is produced by the perception network. The property vector is then utilized by the prediction network to recursively predict future object states $R_1, R_2, \ldots$ for a new rollout given initial state $R_0$. We train the PPN to minimize the L2 distance between the predicted rollout states and the ground truth states for those timesteps.}
    \vspace{-15pt}
    \label{fig:model_arch}
\end{figure*}

\section{RELATED WORK}

Previous methods of modeling intuitive physics have largely fallen under two broad categories: top-down approaches, which infer physical parameters for an existing symbolic physics engine~\cite{ullman2014learning,battaglia2013simulation,bates2015humans,hamrick2011internal,wu2015galileo,wu2016physics}, and bottom-up approaches, which directly predict physical quantities or future motion given observations~\cite{agrawal2016learning,ehrhardt2017learning,fragkiadaki2015learning,lerer2016learning,mottaghi2016newtonian,mottaghi2016happens,sutskever2009recurrent}. While top-down approaches are able to generalize well to any situation supported by their underlying physics engines (\eg different numbers of objects, previously unseen property values, etc.), they are difficult to adapt to situations not supported by their underlying description languages, requiring manual modifications to support new types of interactions. On the other hand, bottom-up approaches are often capable of learning the dynamics of formerly unseen situations without any further modification, though they often lack the ability to generalize in the same manner as top-down approaches.

Recently, a hybrid approach has used neural relation networks, a specific instance of the more general class of graph-based neural networks~\cite{scarselli2009graph,li2015gated}, to attain the generalization benefits of top-down approaches without requiring an underlying physics engine. Relation networks rely on the use of a commutative and associative operation (usually vector addition) to combine pairwise interactions between object state vectors in order to predict future object states~\cite{raposo2017discovering}. These networks have demonstrated success in simulating multiple object dynamics under interactions including Coulomb charge, object collision (with and without perfect elasticity), and spring tension~\cite{battaglia2016interaction,chang2016compositional,watters2017visual,wu2017learning}. Much like a top-down approach, relation networks are able to generalize their predictions of object position and velocity to different numbers of objects (training on 6 objects and testing on 9, for instance) without any modification to the network weights; furthermore, they are fully differentiable architectures that can be trained via gradient descent on a variety of interactions. Our paper leverages the interaction network in a novel way, demonstrating for the first time its efficacy as a perception module and as a building block for unsupervised representation learning. 

Additional research has looked at the supervised and unsupervised learning of latent object properties, attempting to mirror the inference of object properties that humans are able to perform in physical environments~\cite{ullman2014learning}. Wu~\etal~\cite{wu2016physics} leverages a deep model alongside set physical laws to estimate properties such as mass, volume, and material from raw video input. Fraccaro~\etal~\cite{fraccaro2017disentangled} uses a variational autoencoder to derive the latent state of a single bouncing ball domain, which they then simulate using Kalman filtering. Chang~\etal~\cite{chang2016compositional} demonstrate that their relation network based physics simulator is also capable of performing maximum-likelihood inference over a discrete set of possible property values by comparing simulation output for each possibility to reality. Our paper goes one step further by showing that physical properties can be learned from no more than raw motion data of multiple objects. Recently, Kipf~\etal~\cite{kipf2018neural} has also utilized relation networks to infer the identity of categorical interactions between objects; in contrast, our paper is concerned with the learning of object properties. 
\section{MODEL}

\subsection{PERCEPTION-PREDICTION NETWORK}

The PPN observes the physical dynamics of objects with unknown latent properties (\eg mass, coefficient of restitution) and learns to generate meaningful representations of these object properties that can be used for later simulations. An overview of the full network is shown in Figure \ref{fig:model_teaser}. The PPN consists of the following two components: 
\begin{itemize}
    \item The \textbf{perception network} takes as input a sequence of frames on the movements of objects over a short observation window. It outputs a \textit{property vector} for each object in the scene that encodes relevant latent physical properties for that object. Each input frame is a set of \textit{state vectors}, consisting of each object's position and instantaneous velocity. During training, no direct supervision target is given for the property vectors.  
    \item The \textbf{prediction network} uses the property vectors generated by the perception network to simulate the objects from a different starting configuration. The network takes as input the property vectors generated by the perception network and new initial state vectors for all objects. Its output is a rollout of the objects' future states from their new starting state. The training target for the prediction network is the ground truth states of the rollout sequence. 
\end{itemize}

We implement both the perception and prediction networks using interaction networks~\cite{battaglia2016interaction}, a specific type of neural relation network that is fully differentiable and generalizes to arbitrary numbers of objects. This enables us to train both networks end-to-end using gradient descent with just the supervision signal of the prediction network's rollout target, as the property vectors output by the perception network feed directly into the prediction network.

\subsection{INTERACTION NETWORK}

An interaction network (IN) is a relation network that serves as the building block for both the perception and prediction networks. At a high level, interaction networks use multilayer perceptrons (MLPs) to implement two modular functions, the relational model $f_{\text{rel}}$ and the object model $f_{\text{obj}}$, which are used to transform a set of object-specific input features $\{x^{(1)}, \ldots, x^{(N)}\}$ into a set of object-specific output features $\{y^{(1)}, \ldots, y^{(N)}\}$, where $N$ is the number of objects in a system. Given input features for two objects $i$ and $j$, $f_{\text{rel}}$ calculates the ``effect" vector of object $j$ on object $i$ as $e^{(i,j)} = f_{\text{rel}}(x^{(i)}, x^{(j)})$. The net effect on object $i$, $e^{(i)}$, is the vector sum of all pairwise effects $\sum_{j \neq i} e^{(i,j)}$ on object $i$. Finally, the output for object $i$ is given by $y^{(i)} = f_{\text{obj}}(x^{(i)}, e^{(i)})$. Importantly, $f_{\text{obj}}$ and $f_{\text{rel}}$ are shared functions that are applied over all objects and object-object interactions, allowing the network to generalize across variable numbers of objects.

Interaction networks are capable of learning state-to-state transition functions for systems with complex physical dynamics. More generally, however, interaction networks can be used to model functions where input and output features are specific to particular objects and the relationship between input and output is the same for each object. While our prediction network uses an interaction network to simulate state transitions, our perception network uses an interaction network to make incremental updates on the values of object latent properties from observed evidence. 

\subsection{PERCEPTION NETWORK}

The perception network produces object-specific property vectors, $Z$, from a sequence of observed states $O$. As shown in Figure \ref{fig:model_arch}, our perception network is a recurrent neural network that uses an interaction network as its core recurrent unit. The perception network begins with object-specific code vectors, $C_1$, initialized to zero vectors, with some fixed size $L_C$ for each object. At each step $t$, the IN takes in the previous code vectors, $C_{t-1}$, as well as the last two observed states, $O_{t-1}$ and $O_t$, to produce updated code vectors, $C_t$, also of size $L_C$. After processing all $T_O$ observation frames, the perception network feeds the final code vectors $C_{T_O}$ into a single code-to-property MLP that converts each object's code vector into an ``uncentered" property vector of size $L_Z$ per object. We denote the final collection of uncentered property vectors as $Z_u$.

In many physical systems, it may be impossible or undesirable to measure the latent properties of objects on an absolute scale. For example, in a system where two balls collide elastically, a collision can only inform us on the mass of each object relative to the other object, not their absolute mass values. In order to allow for the inference of absolute property values, we let the first object of every system serve as a \textit{reference object} and take on the same property values in each system. In doing so, we can infer the absolute property values of all other objects by observing their value relative to the reference object. To enforce inference relative to the reference object, we ``center" the property vectors by subtracting the reference object's uncentered property vector from each object's uncentered property vector, producing the final property vectors $Z$. Note that this ensures that the reference object's property vector is always a zero vector, agreeing with the fact that its properties are known to be constant. We can summarize the perception network with the following formulas: 
\begin{align}
    C_1 &= \boldsymbol{0} \\
    C_t &= \text{IN}_{pe}(C_{t-1} \Vert O_{t-1} \Vert O_t),  \text{ for } t=2,\ldots,T_O \\
    Z_u^{(i)} &= \text{MLP}_{pe}\left(C_{T_O}^{(i)}\right), \text{ for } i=1,\ldots,N \\
    Z^{(i)} &= Z_u^{(i)} - Z_u^{(1)}, \text{ for } i=1,\ldots,N
\end{align}
where $\Vert$ is the object-wise concatenation operator, $\text{IN}_{pe}$ is the perception interaction network, $\text{MLP}_{pe}$ is the code-to-property MLP, and $Z_u^{(1)}$ is the reference object's uncentered property vector.

\subsection{PREDICTION NETWORK}

The prediction network performs state-to-state rollouts of the system from a new initial state, $R_0$, using the property vectors produced by the perception network. Like the perception network, the prediction network is a recurrent neural network with an Interaction Network core. At step $t$, the IN takes in the previous state vectors, $R_{t-1}$, and the property vectors, $Z$, and outputs a prediction of the next state vectors, $R_{t}$. In other words, 
\begin{equation}
    R_{t} = \text{IN}_{pr}(R_{t-1} \Vert Z), \text { for } t=1,...,T_R
\end{equation}
where $\text{IN}_{pr}$ is the prediction interaction network and $T_R$ is the number of rollout frames. 

The prediction loss for the model is the total MSE between the predicted and true values of $\{R_t\}_{t=1...T_R}$.

\section{EXPERIMENTS}
\subsection{PHYSICAL SYSTEMS}

For our experiments, we focus on 2-D domains where both the latent property inference task and the subsequent dynamics prediction task are challenging. In all systems, the first object serves as the reference object and has fixed properties. All other objects' properties can be inferred relative to the reference object's properties. We evaluate the PPN on the following domains (see Fig. \ref{fig:ro}): 
\begin{itemize}
    \myitem \textbf{Springs} Balls of equal mass have a fictitious property called ``spring charge" and interact as if all pairs of objects were connected by springs governed by Hooke's law\footnote{Two objects connected by a spring governed by Hooke's law are subject to a force $F = -k (x - x_0)$, where $k$ is the spring constant of the spring, $x$ is the distance between the two objects, and $x_0$ is the spring's equilibrium distance. The force is directed along the line connecting the two objects but varies in sign: it is attractive if $x > x_0$ and repulsive if $x < x_0$.}. The reference object has a spring charge of 1, while all other objects have spring charges selected independently at random from the log-uniform\footnote{We use the phrase log-uniform distribution over $[A, B]$ to indicate the distribution of $\exp(x)$, where $x$ is drawn uniformly at random over the interval $[\log A, \log B]$.} distribution over $[0.25, 4]$. The spring constant of the spring connecting any given pair of objects is the product of the spring charges of the two objects, and the equilibrium distance for all springs is a fixed constant.
    \myitem \textbf{Perfectly Elastic Bouncing Balls} Balls of fixed radius bounce off each other elastically in a closed box. The reference object has a mass of 1. Each other ball has a mass selected independently at random from the log-uniform distribution over $[0.25, 4]$. The four walls surrounding the balls have infinite mass and do not move. 
    \myitem \textbf{Inelastic Bouncing Balls} Building off the previous domain, we introduce additional complexity by adding coefficient of restitution (COR) as another varying latent property of each object. The COR of a collision is the ratio of the final to initial relative velocity between the two colliding objects along the axis perpendicular to the contact plane. In a perfectly elastic domain, for example, all collisions would have a COR of 1. In our new domain, each object has a random COR selected uniformly from $[0.5, 1]$. The reference object has a COR of $0.75$. The COR used to compute the dynamics in a collision between two balls is defined as the maximum of the two colliding objects' CORs. When a ball collides with a wall, the ball's COR is used for the collision.
\end{itemize}

For each domain, we train the PPN on a 6-object dataset with $10^6$ samples and validate on a 6-object dataset with $10^5$ samples. Each sample consists of 50 observation frames used as input into the perception network and 24 rollout frames used as targets by the prediction network. We evaluated our model on 3-object, 6-object, and 9-object test sets, each with $10^5$ samples.

In addition, we also wish to demonstrate the PPN's ability to generalize to new objects whose latent properties are outside of the range of values seen during training. For this experiment, we test our model on a new 2-object perfectly elastic balls dataset with $10^5$ samples. The mass of the first ball remains fixed at 1, while the mass of the second ball is selected from 11 values ranging from $32^{-1}$ to $32$, spaced evenly on a log scale. We perform a similar experiment on the springs domain, using the same 11 values as the spring charge of the second object.

We use matter-js\footnote{\url{http://brm.io/matter-js/}}, a general-purpose rigid-body physics engine, to generate ground truth data. In all simulations, balls are contained in a $512$ px $\times$ $512$ px closed box. Each ball has a 50 px radius and randomly initialized positions such that no ball overlaps. In the springs domain, initial x- and y-velocity components are selected uniformly at random from the range $[-15, 15]$ px/sec, the equilibrium displacement for each spring is 150, and the mass of all balls is $10^4$. In the perfectly elastic balls domain, initial velocity components are selected from the range $[-9, 9]$ px/sec. In the inelastic balls domain, they are selected from the range $[-13, 13]$ px/sec. Each dataset's frames are sampled at $120$ fps. 

In the creation of our bouncing ball datasets, we use rejection sampling to filter out simulations in which some object latent properties cannot be inferred from the observation frames. In both bouncing ball domains, we must be able to infer the mass of every object. In order to guarantee this, each object must collide directly with the reference object or be linked indirectly to it through a sequence of collisions. For the inelastic domain, we must ensure that each object's COR can be inferred as well. In a ball-ball collision, only the higher object COR is used in determining collision dynamics, and so only the higher object COR can be inferred from the collision. For this reason, every ball must either collide with a ball of lower COR or a wall.  

\subsection{MODEL ARCHITECTURE}

\begin{table*}[t]
    \centering
    \setlength{\tabcolsep}{5pt}
    \begin{tabular}{c c c c c c c c c c}
        \toprule
        & 
        \multicolumn{2}{c}{Springs} & & \multicolumn{2}{c}{Perfectly Elastic Balls} & & \multicolumn{3}{c}{Inelastic Balls} 
        \\
        \cmidrule(lr){2-3}
        \cmidrule(lr){5-6} 
        \cmidrule{8-10}
        Component \# & 
        EVR & $R^2$ w/ log charge & & 
        EVR & $R^2$ w/ log mass & & 
        EVR & $R^2$ w/ log mass & $R^2$ w/ COR
        \\
        \midrule
        1 & 0.94 & 0.95 && 0.99 & 0.94 && 0.73 & 0.90 & 0.02 \\
        2 & 0.06 & 0.02 && 0.006 & 0 && 0.27 & 0.02 & 0.81 \\
        3 & 0 & 0 && 0 & 0 && 0.006 & 0 & 0 \\
        4 & 0 & 0 && 0 & 0 && 0 & 0 & 0 \\
        \bottomrule
    \end{tabular}
    \vspace{-7pt}
    \caption{\textbf{Principal component analysis.} Applying PCA on the property vectors yields principal components that are highly correlated with human-interpretable latent properties such as COR and the log of mass. We compute statistics on the first four principal components of the property vectors for each training set. Explained variance ratio or EVR is the explained variance of the principal component as a fraction of overall variance, and $R^2$ is the squared in-sample correlation between the principal component and a particular ground truth property. Values less than $10^{-3}$ round to 0.}
    \vspace{-5pt}
    \label{tab:var}
\end{table*}
\begin{table*}[t]
    \centering
        \begin{tabular}{cccccc}
        \toprule
        \multirow{2}{*}{\# Training Data} & \multirow{2}{*}{\# Test Objects} &
        Springs & Perfectly Elastic Balls & \multicolumn{2}{c}{Inelastic Balls}
        \\
        \cmidrule(lr){3-3}\cmidrule(lr){4-4}\cmidrule(lr){5-6}
        & & $R^2$ w/ log charge & $R^2$ w/ log mass & $R^2$ w/ log mass & $R^2$ w/ COR\\
        \midrule
        10$^5$ & 6 & 0.60 & 0.91 & 0.55 & 0.03 \\
        $2\times10^5$ & 6 & 0.95 & 0.96 & 0.95 & 0.65 \\
        $5\times10^5$ & 6 & 0.94 & 0.94 & 0.91 & 0.77 \\
        10$^6$ & 6 & 0.95 & 0.94 & 0.90 & 0.80 \\
        \midrule
        \multirow{2}{*}{10$^6$} & 3 & 0.90 & 0.97 & 0.92 & 0.86 \\
        & 9 & 0.87 & 0.92 & 0.90 & 0.68 \\
        \bottomrule
    \end{tabular}
    \vspace{-5pt}
    \caption{\textbf{Data-efficiency and number of objects generalization.} The PPN learns to capture physical properties with 10$^5$ training data points and converges when given $2\times10^5$ instances. Its predictions generalize well to out-of-sample test sets with varying numbers of objects. We train the PPN on a 6-object dataset and test it on entirely new datasets comprised of 6, 3, and 9 objects. Above, we report the $R^2$ when using the property vector's first principal component to predict log mass and the second principal component to predict COR (for the inelastic balls case). Note that even in the 3 and 9 object cases the PPN is able to extract mass and coefficient of restitution with high $R^2$.}
    \vspace{-13pt}
    \label{tab:oosr2}
\end{table*}
\begin{figure*}[t]
    \centering
    \includegraphics[width=0.45\linewidth]{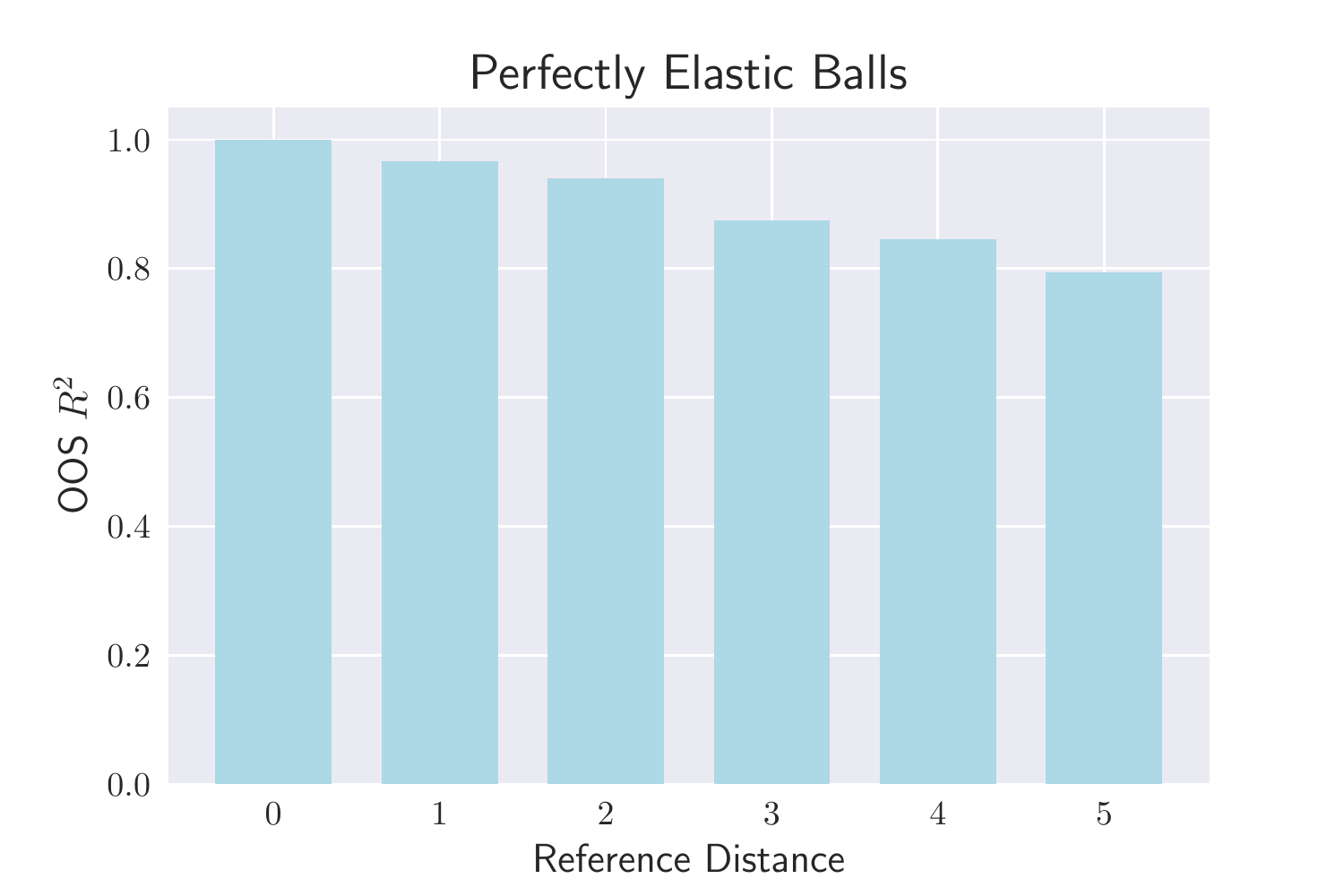}
    \includegraphics[width=0.45\linewidth]{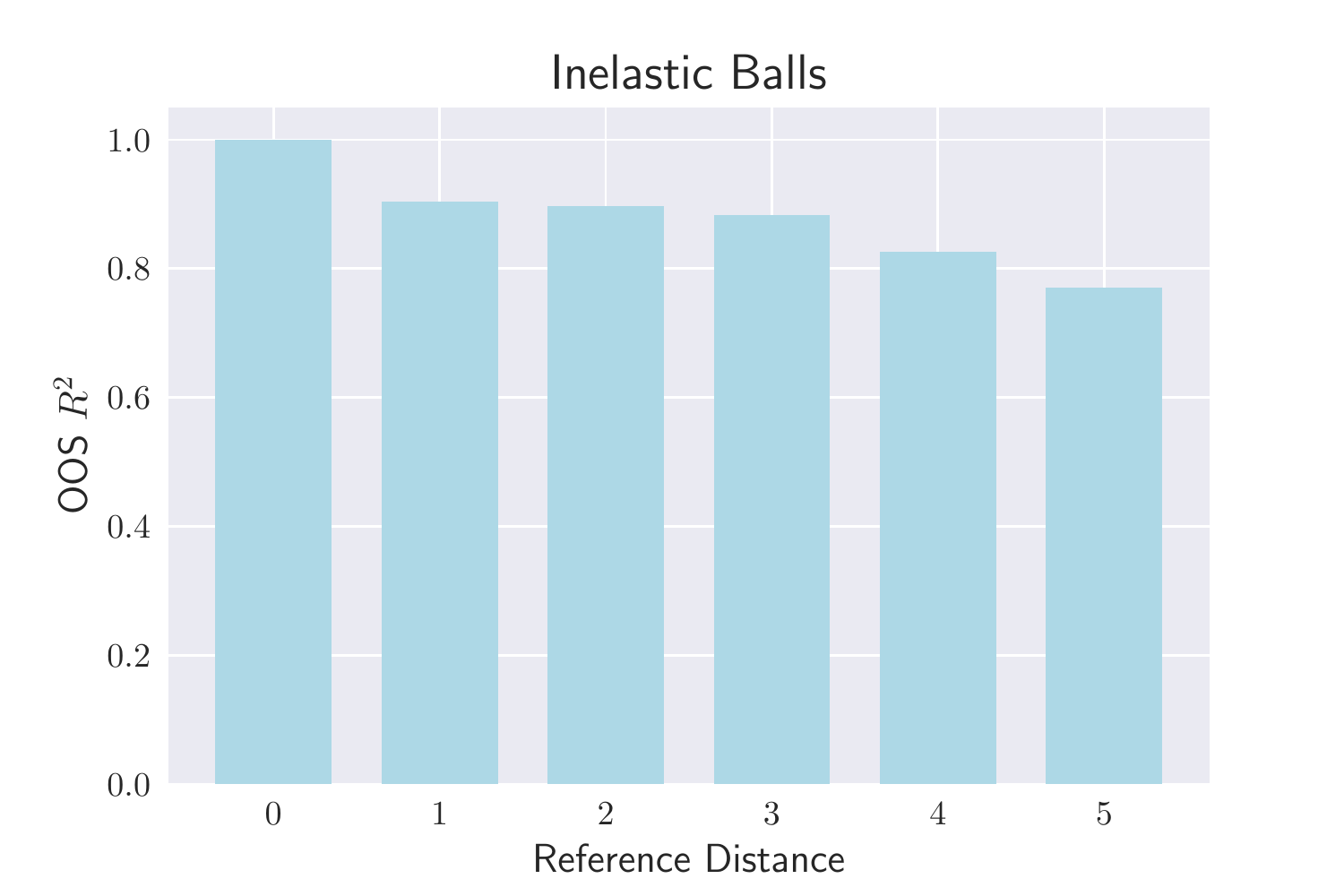}
    \vspace{-10pt}
    \caption{\textbf{Mass prediction vs. reference distance.} Out-of-sample $R^2$ on the two 6-object bouncing balls datasets for predicting log mass at different reference distances. The PPN must combine a sequence of intermediate mass inferences to accurately infer the mass of an object with large reference distance.}
    \label{fig:ref_dist}
    \vspace{-15pt}
\end{figure*}

We use a single model architecture for all of our experiments. We set $L_C$, the size of each code vector, to 25 and $L_Z$, the size of each property vector, to 15. All MLPs in the model, including those in the interaction networks, use linear hidden layers with ReLU activation and a linear output layer. 

Following the overall structure of Battaglia~\etal~\cite{battaglia2016interaction}, the perception network's IN core consists of a 4-layer relation-centric MLP with sizes $[75, 75, 75, 50]$ and a 3-layer object-centric MLP with sizes $[50, 50, 25]$. The final code vectors output by the IN feed into another object-centric MLP of size $[15, 15, 15]$ to produce the final latent property vectors of size $15$. The prediction network's IN core consists of a 5-layer relation-centric MLP with sizes $[100, 100, 100, 100, 50]$ and a 3-layer object-centric MLP with sizes $[50, 50, 4]$ used to predict each object's next position and velocity. 

The perception network and prediction network are trained end-to-end using a single training loss, which we call the prediction loss. The prediction loss is the unweighted sum of the MSE of the predicted vs actual state vectors of all objects during the 24 rollout timesteps. In addition, we apply L2 regularization on the ``effects" layer of both the perception and prediction networks. This regularization encourages minimal information exchange during interactions and proves to be a crucial component to generalization to different numbers of objects. We selected the penalty factor for each regularization term via grid search. We also experimented with the use of $\beta$-VAE regularization~\cite{kingma2013auto,higgins2016beta} on property vectors to encourage the learning of interpretable and factorized properties. 

In order to improve stability when simulating long rollouts, we added a small amount of Gaussian noise to each state vector during rollout, forcing the model to self-correct for errors. Empirically, we found that setting the noise std. dev. equal to $0.001\times$ the std. dev. of each state vector element's values across the dataset stabilized rollout positions without affecting loss.

We trained the model for $150$ epochs and optimized the parameters using Adam~\cite{kingma2014adam} with mini-batch size 256. We used a waterfall schedule that began with a learning rate of $5 \times 10^{-4}$ and downscaled by $0.8$ each time the validation error, estimated over a window of $10$ epochs, stopped decreasing.

\section{RESULTS}

\subsection{EXTRACTING LATENT PROPERTIES}

Our results show that the physical properties of objects are successfully encoded in the property vectors output by the perception network. In fact, we can extract the human-interpretable notions of spring charge, mass, and COR by applying principal component analysis (PCA) to the property vectors generated by the perception network during training. We find that the first principal component of each property vector is highly correlated with the log of spring charge in the spring domain and the log of object mass in both bouncing ball domains. In the inelastic balls domain, we also find that the second principal component of the property vector is highly correlated with COR. Table \ref{tab:var} shows the explained variance ratio (EVR) of each of the first 4 principal components of the learned property vectors in all three domains, along with the $R^2$ when each component is used to predict ground truth object properties\footnote{By default, the property values produced by PCA will not be in the same scale as our ground truth values. For the purposes of correlation analysis, we linearly scale predictions to match the mean and std. dev. of the ground truth latent values.}. Since PCA is an unsupervised technique, these scalar quantities can be discovered without prior notions of mass and COR, and we can use the order-of-magnitude difference between certain principal components' EVR to identify which components represent meaningful properties and which merely capture noise.

We also find that each learned property vector only contains information about its associated object and not any other objects. We test this hypothesis by using linear least squares to calculate the in-sample $R^2$ between the ground truth latent properties of each object and the concatenation of the property vectors of all other objects. This $R^2$ is less than 5\% for each of the three domains and their relevant latent properties. 

In order to test the generalization properties of our perception network, we calculate the out-of-sample $R^2$ when using the perception network (trained on 6 object dynamics) and PCA to predict property values for test sets with varying number of objects, as shown in Table \ref{tab:oosr2}. The table also shows how PPN performs when given a different number of training instances. In all bouncing balls test sets, for our model trained on $10^6$ data points, the OOS $R^2$ for log mass is above 90\%, the OOS $R^2$ for COR is above 68\%, and the OOS $R^2$ for log spring charge in the springs domain is above 87\%.

We also compare the PPN against a \textbf{LSTM-PPN} baseline. The LSTM-PPN replaces each of the perception and prediction networks in the PPN with stacked LSTMs. Unlike an interaction network, an LSTM does not factorize input and output by object. Instead, state vectors for each object are concatenated and processed together, and a single property vector is learned for all objects. Table \ref{tab:cmpr} shows that the LSTM-PPN does not learn meaningful latent properties. In each scenario, the linear least squares in-sample $R^2$ between true object properties and property vectors is less than $2\%$. We also experiment with different values of $\beta$ in the regularization term of the property vectors $Z$ as in $\beta$-VAE~\cite{higgins2016beta}. The value of $\beta$ does not impact the PPN's performance on learning object properties.

For the two bouncing balls domains, the relative masses of objects are inferred through collisions, but not all objects collide directly with the reference object. We define the \textit{reference distance} of an object to be the minimum number of collisions needed during observation to relate the object's mass to that of the reference object. Inference on an object with reference distance of 3, for example, depends on the inference of the mass of two intermediate objects. Figure \ref{fig:ref_dist} shows the relation between the PPN's prediction $R^2$ and reference distance for each of the 6-object test sets. While there is a decay in $R^2$ as reference distance increases due to compounding errors during inference, the PPN clearly demonstrates the ability to use transitivity to infer the mass of objects with large reference distance.

\begin{table}[t]
    \centering\small
    \setlength{\tabcolsep}{3pt}
        \begin{tabular}{lcccc}
        \toprule
        \multirow{2}{*}{Methods} & Springs & Elastic Balls & \multicolumn{2}{c}{Inelastic Balls}
        \\
        \cmidrule(lr){2-2}\cmidrule(lr){3-3}\cmidrule(lr){4-5}
        & log charge & log mass & log mass & COR\\
        \midrule
        LSTM & 0.02 & 0.03 & 0.02 & 0.03\\
        PPN ($\beta=0$) & {\bf 0.95} & {\bf 0.94} & 0.90 & {\bf 0.80} \\
        PPN ($\beta=0.01$) & {\bf 0.95} & 0.93 & {\bf 0.93} & 0.79\\
        PPN ($\beta=1$) & 0.92 & {\bf 0.94} & {\bf 0.93} & 0.65\\
        \bottomrule
    \end{tabular}
    \vspace{-5pt}
    \caption{\textbf{Comparing with baseline methods}. Varying the value of $\beta$ in the regularization term as in $\beta$-VAE does not change the PPN's performance significantly. The PPN consistently outperforms the baseline LSTM.}
    \vspace{-15pt}
    \label{tab:cmpr}
\end{table}

\begin{figure*}[t]
    \centering
    \begin{tabular}{c c c c c c c c c c c c}
        \toprule
        & \multicolumn{3}{c}{Springs}
        && \multicolumn{3}{c}{Perfectly Elastic Balls}
        && \multicolumn{3}{c}{Inelastic Balls} 
        \\
        \cmidrule(lr){2-4}
        \cmidrule(lr){6-8}
        \cmidrule(lr){10-12}
        Model 
        & 6 balls & 3 balls & 9 balls 
        && 6 balls & 3 balls & 9 balls 
        && 6 balls & 3 balls & 9 balls
        \\
        \midrule
        PPN 
        & 0.020 & 0.078 & 0.057
        && 0.025 & 0.017 & 0.032 
        && 0.048 & 0.041 & 0.054 
        \\
        MPPR
        & 0.124 & 0.082 & 0.139
        && 0.038 & 0.027 & 0.046
        && 0.062 & 0.045 & 0.073
        \\
        GPIN
        & 0.005 & 0.068 & 0.043 
        && 0.019 & 0.015 & 0.027 
        && 0.029 & 0.021 & 0.039 
        \\
        \bottomrule
    \end{tabular}
    
    \includegraphics[width=0.33\linewidth]{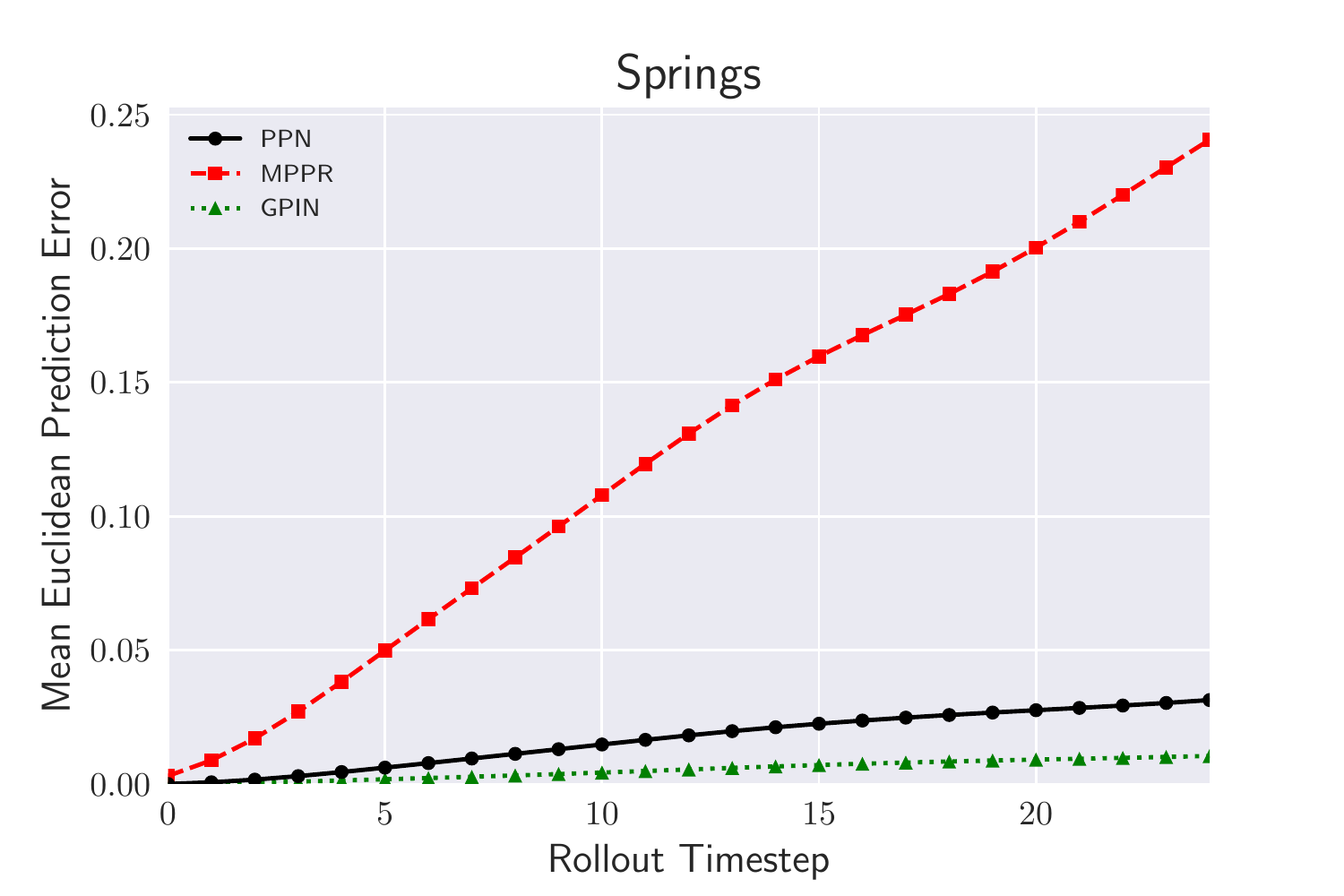}
    \includegraphics[width=0.33\linewidth]{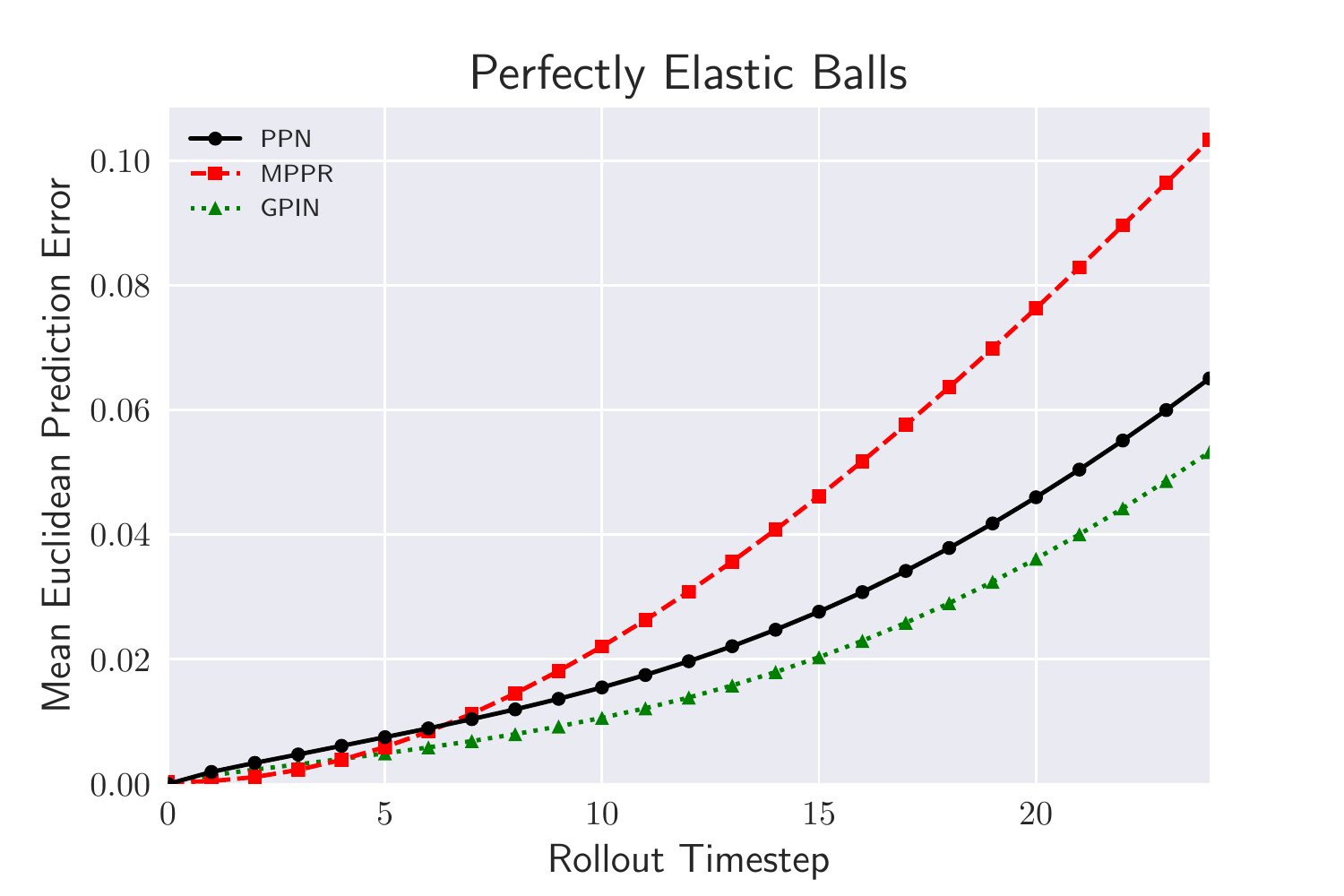}
    \hfill
    \includegraphics[width=0.33\linewidth]{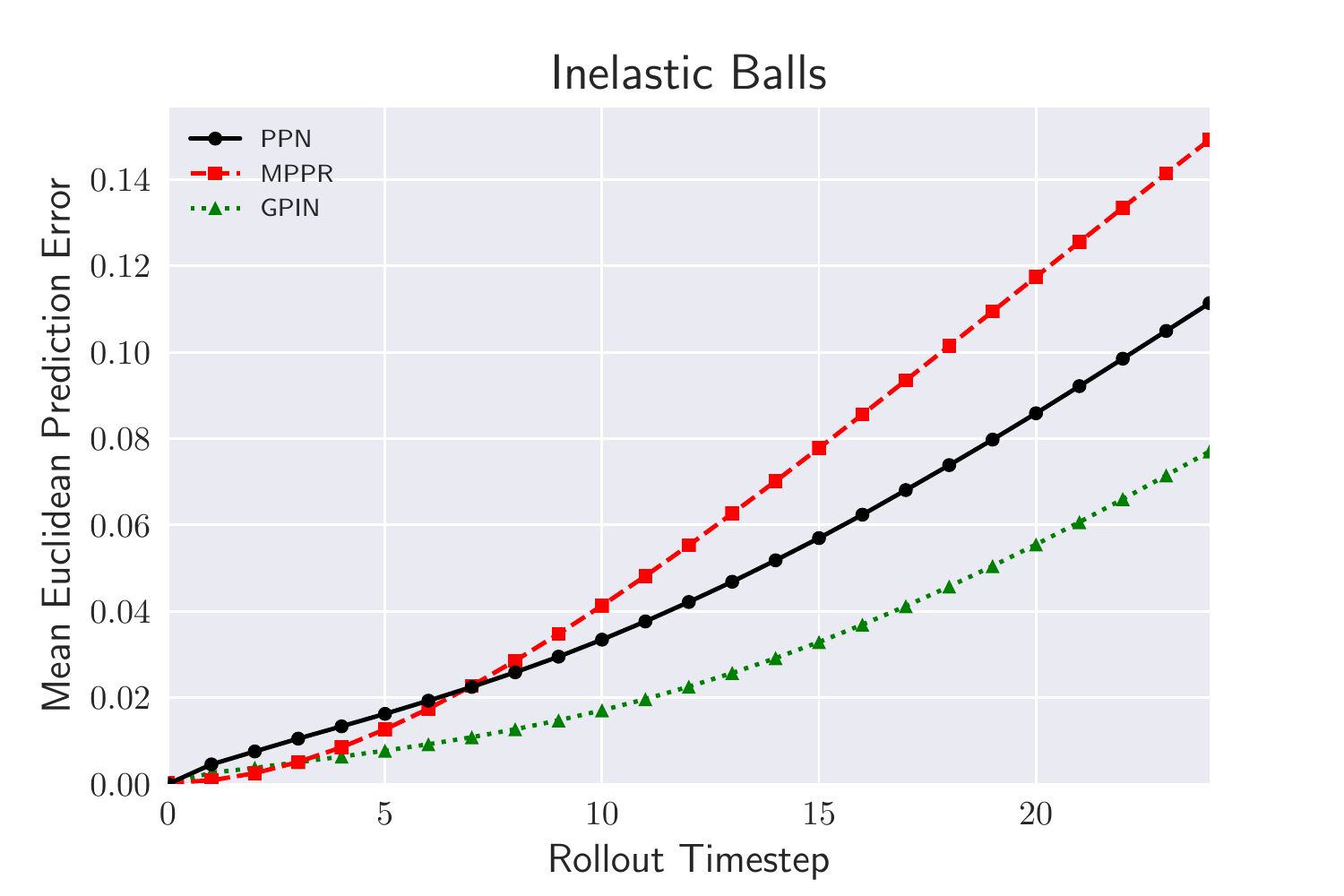}
    
    \vspace{-10pt}
    \caption{\textbf{Mean Euclidean prediction error.} \textit{Top:} Mean Euclidean prediction error over all timesteps and samples for each test set measured as fraction of framewidth. For each domain, the PPN and GPIN are trained on 6-object systems and tested on new systems with 6, 3, and 9 objects. \textit{Bottom:} Mean Euclidean prediction error at different rollout timesteps for each of the 6-object scenarios. Plots for the 3-object and 9-object scenarios exhibit similar behavior.}
    \label{fig:pred}
    \vspace{-10pt}
\end{figure*}

\def \figscale {0.13}
\begin{figure*}[t!]
    \centering
    \setlength{\tabcolsep}{1pt}
    \fontfamily{phv}\fontsize{9}{11}\selectfont
    \begin{tabular}{c c c @{\hskip 0.2cm} c c @{\hskip 0.2cm} c c}
        &
        \multicolumn{2}{c}{6 Balls}
        &
        \multicolumn{2}{c}{3 Balls}
        &
        \multicolumn{2}{c}{9 Balls}
        \\
        & True & Model & True & Model & True & Model
        \\
         \rotatebox{90}{\hspace{0.55cm}Springs}
        &
        \includegraphics[scale=\figscale]{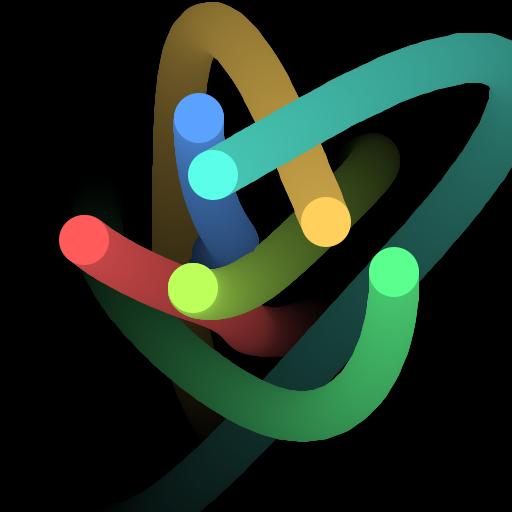}
        &
        \includegraphics[scale=\figscale]{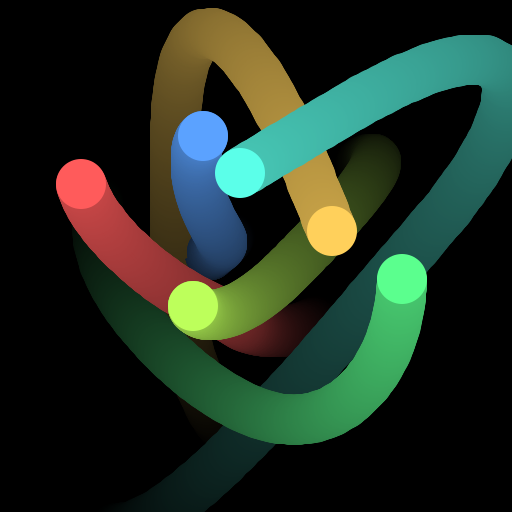}
        &
        \includegraphics[scale=\figscale]{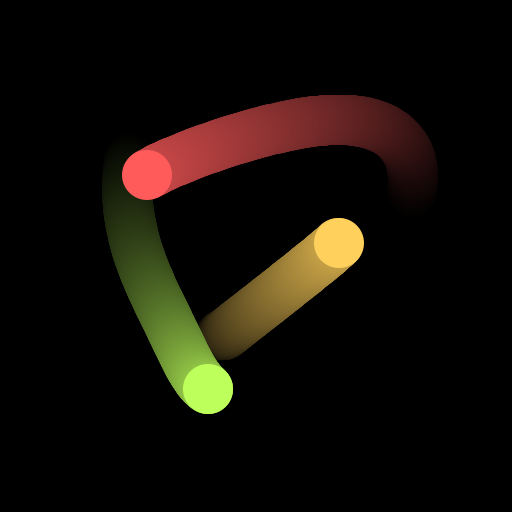}
        &
        \includegraphics[scale=\figscale]{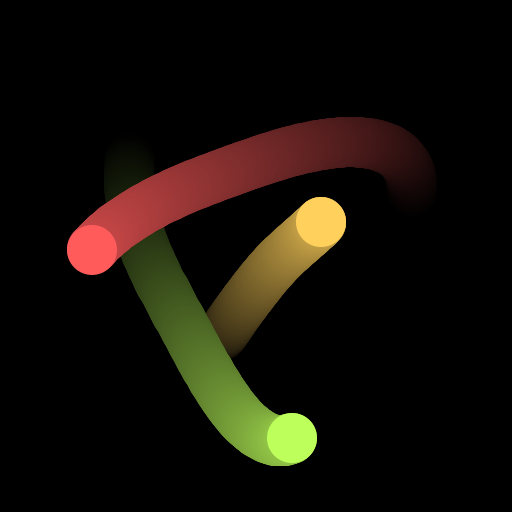}
        &
        \includegraphics[scale=\figscale]{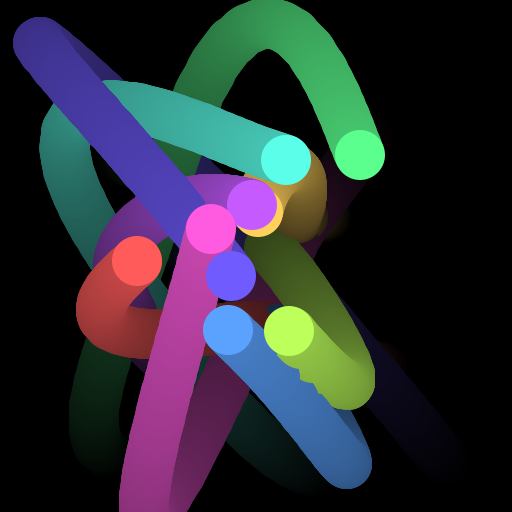}
        &
        \includegraphics[scale=\figscale]{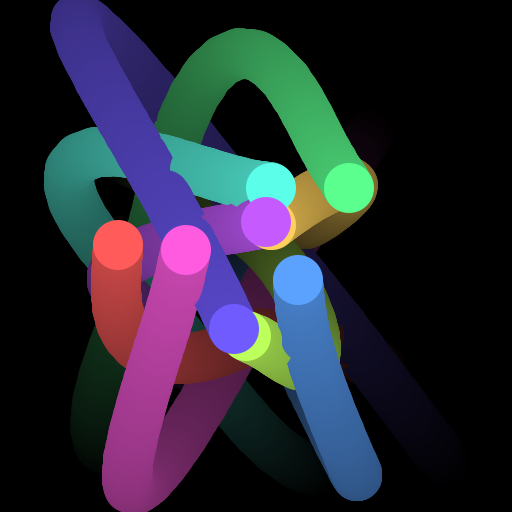}
        \\
        \rotatebox{90}{\hspace{0.05cm}Perfectly Elastic}
        &
        \includegraphics[scale=\figscale]{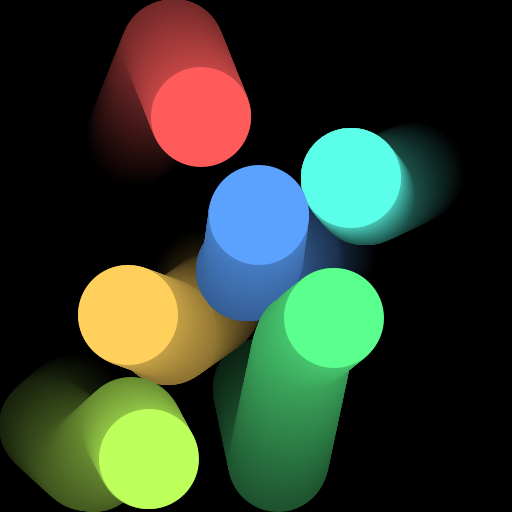}
        &
        \includegraphics[scale=\figscale]{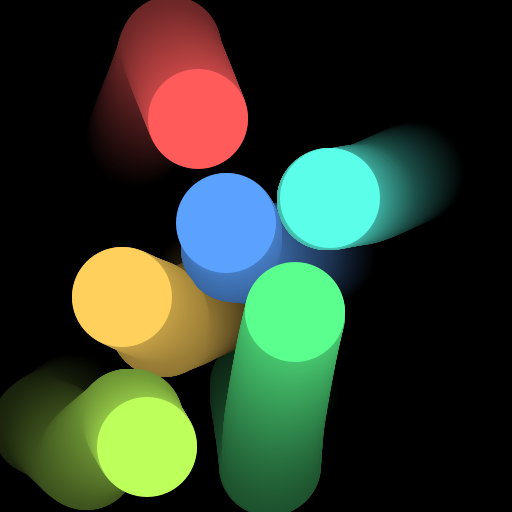}
        &
        \includegraphics[scale=\figscale]{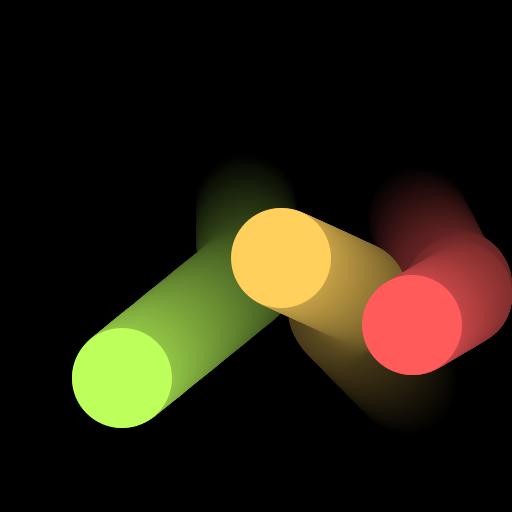}
        &
        \includegraphics[scale=\figscale]{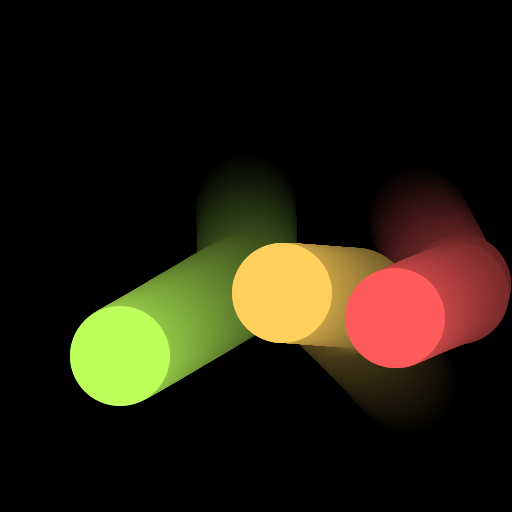}
        &
        \includegraphics[scale=\figscale]{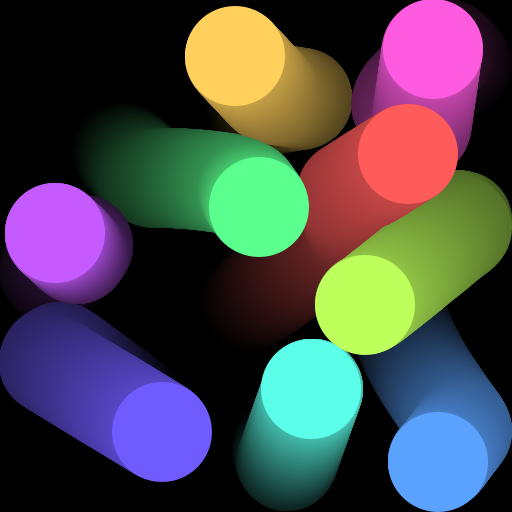}
        &
        \includegraphics[scale=\figscale]{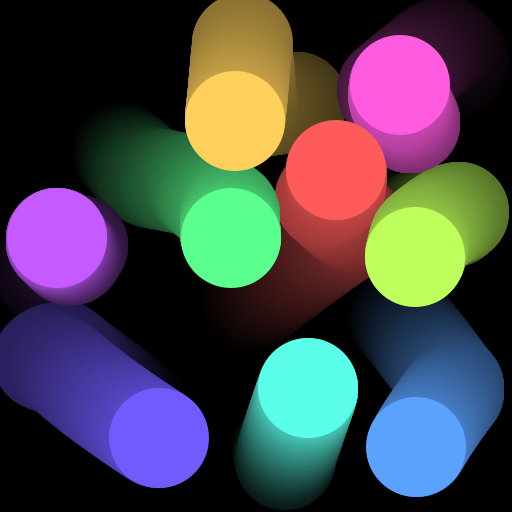}
        \\
        \rotatebox{90}{\hspace{0.6cm}Inelastic}
        &
        \includegraphics[scale=\figscale]{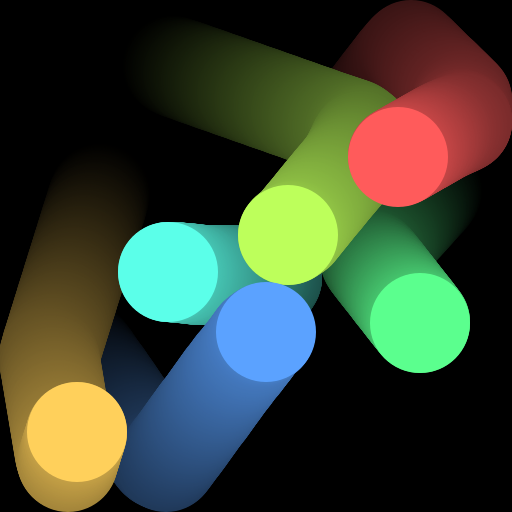}
        &
        \includegraphics[scale=\figscale]{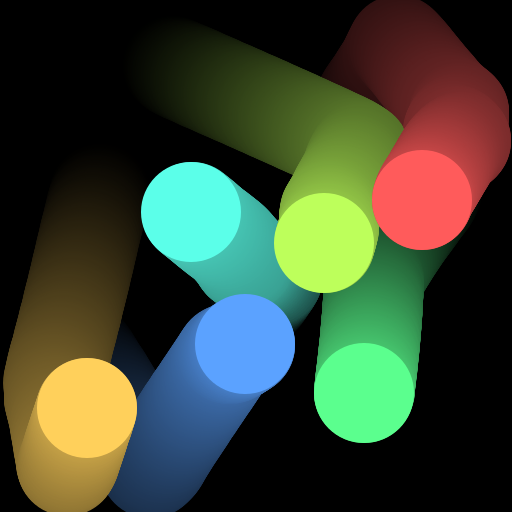}
        &
        \includegraphics[scale=\figscale]{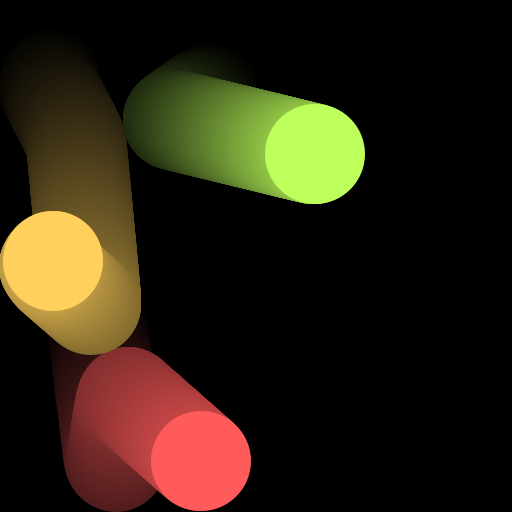}
        &
        \includegraphics[scale=\figscale]{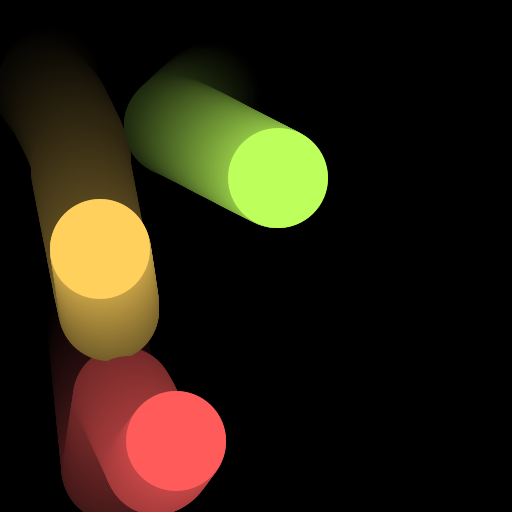}
        &
        \includegraphics[scale=\figscale]{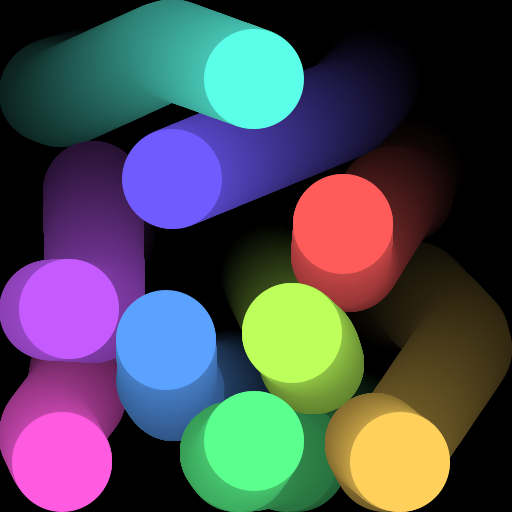}
        &
        \includegraphics[scale=\figscale]{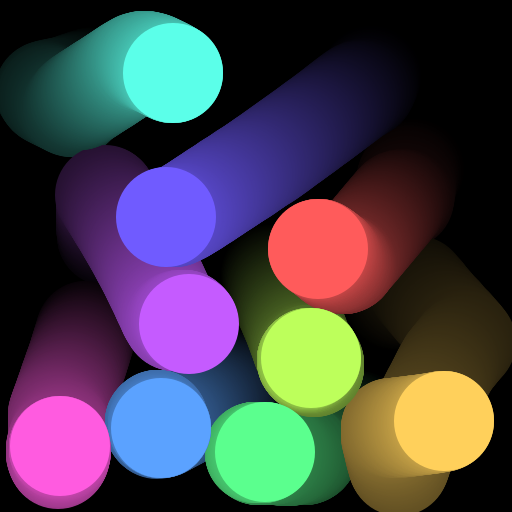}
        \\
    \end{tabular}
    \vspace{-8pt}
    \caption{\textbf{Rollout trajectories.} Sample rollout trajectories (over 24 timesteps) from each of the six test sets. Each domain's model was trained on 6-object samples and tested on 6-, 3-, and 9-object samples.}
    \label{fig:ro}
    \vspace{-15pt}
\end{figure*}

\begin{figure*}[t]
    \centering
    \includegraphics[width=0.45\linewidth]{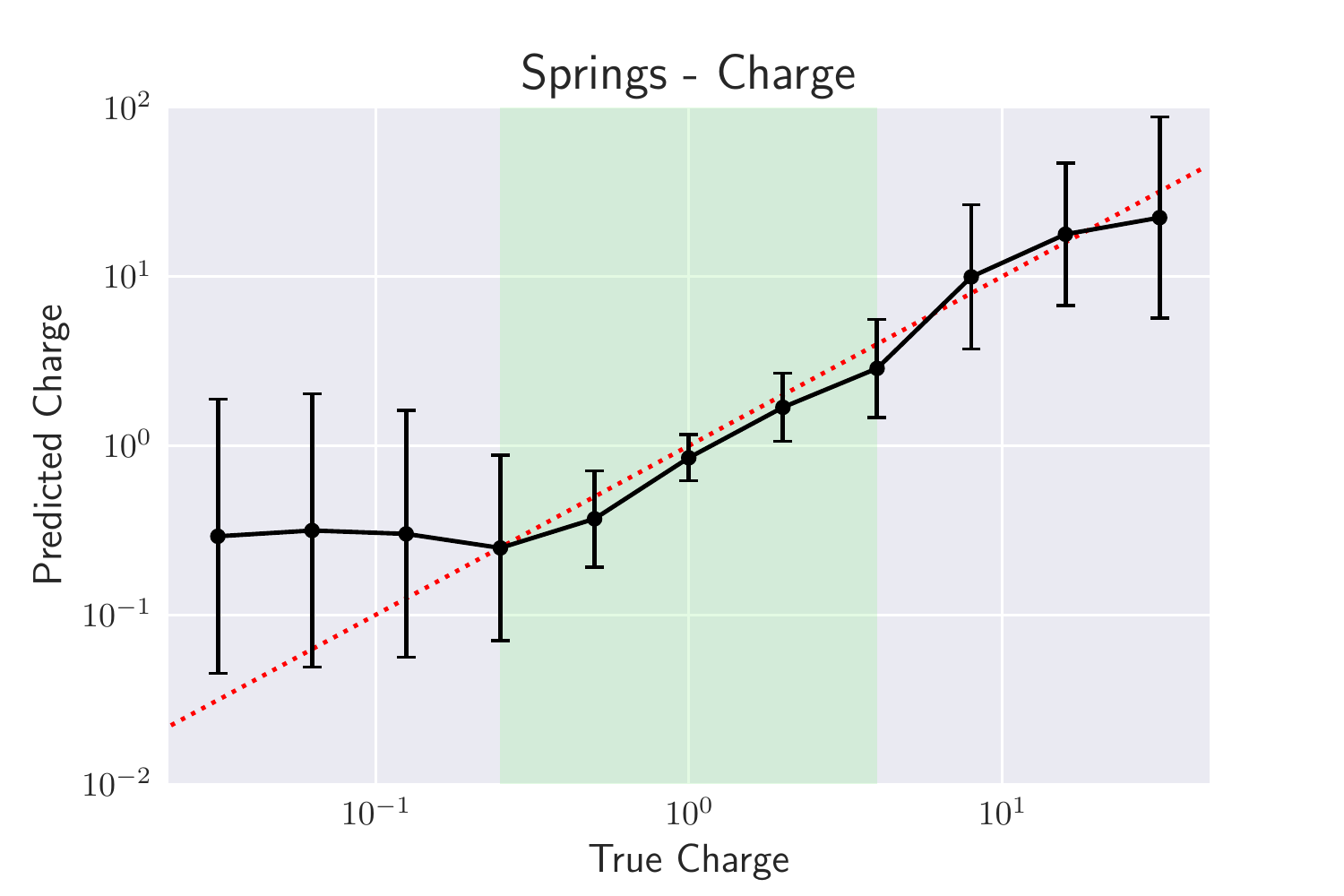}
    \includegraphics[width=0.45\linewidth]{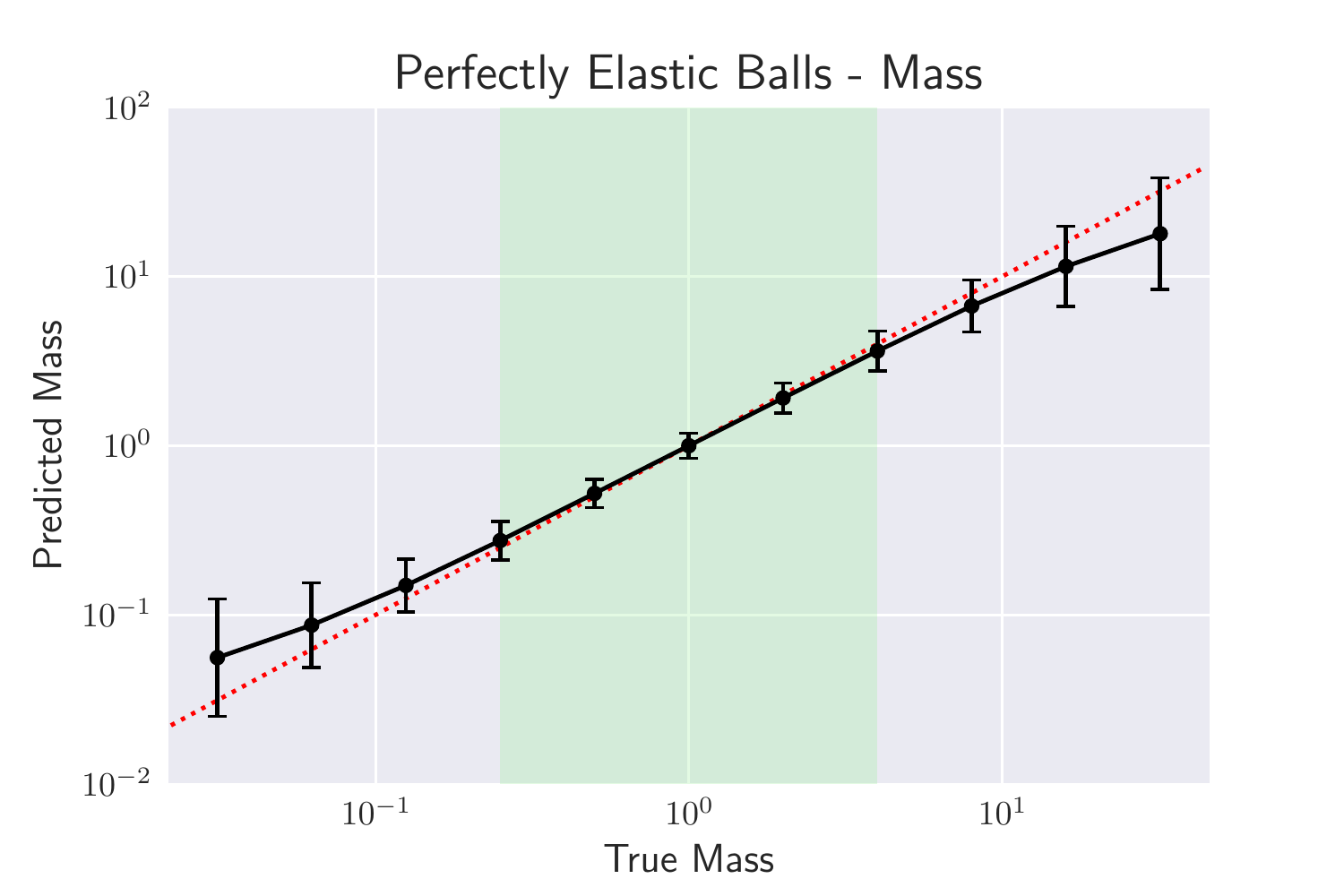}
    \vspace{-8pt}
    \caption{\textbf{Property value generalization.} Predicted property values vs. true property values of the second object in the 2-object test sets for both the springs and perfectly elastic balls domains. The true property values range from $32^{-1}$ to $32$, and the green region, $4^{-1}$ to $4$, indicates property values which appear to the PPN during training. Error bars show 95\% confidence intervals. On the whole, the PPN continues to make reasonable predictions on mass and spring charge values well outside the training set, though the prediction of objects with lower spring charge than previously encountered is noticeably worse.}
    \vspace{-15pt}
    \label{fig:mass}
\end{figure*}

\subsection{ROLLOUT PREDICTIONS}

Although the PPN's primary objective is the unsupervised learning of latent physical properties, the network can also be used to simulate object dynamics. To evaluate the PPN's prediction performance, we use the mean Euclidean prediction error, or the mean Euclidean norm between the ground truth and predicted rollout positions, averaged over all samples and objects. We compare the PPN's performance against two benchmarks. The \textbf{Mean Properties Perfect Rollout (MPPR)} baseline outputs a perfect rollout from the starting state, but incorrectly assumes that all object masses and spring charges are 1. For the inelastic balls domain, it also assumes that all object CORs are 0.75. The \textbf{Ground Truth Properties Interaction Network (GPIN)} benchmark is an IN with the same architecture as the PPN's prediction network. Unlike the PPN, it has direct access to ground truth latent values as input, though it is still only trained on 6-object datasets. Figure \ref{fig:pred} lists the three models' mean Euclidean prediction errors for various scenes and shows how the prediction errors vary for different rollout steps. The PPN's mean Euclidean prediction error is significantly better than the MPPR baseline and comes reasonably close to the GPIN model, especially for the springs and perfectly elastic balls datasets. 

Finally, Figure \ref{fig:ro} shows visualizations of the PPN's rollout trajectories. Randomly selected simulations can be found at \url{http://ppn.csail.mit.edu}. Like the original IN, the PPN's rollouts are sensitive to small prediction errors in early timesteps, but remain visually convincing. 

\subsection{GENERALIZING TO NEW OBJECTS}

Our experiments also explore generalizations to objects whose property values are outside the range found in the training set. We test the PPN framework on a 2-object perfectly elastic test set where the second ball's mass varies from $32^{-1}$ to $32$. Mass values in the range $[0.25, 4]$ are found within the training set, while mass values outside this range require the PPN to extrapolate its understanding of mass to values it has not previously been exposed to. We perform a similar experiment on the springs domain, in which the second object's spring charge varies from $32^{-1}$ to $32$. Figure \ref{fig:mass} plots the relationship between true and predicted property values for the second ball in the two domains, using the same PCA technique described in Section 5.1 to make predictions. 

In the perfectly elastic balls domain, the PPN continues to offer accurate predictions of mass even when the true value lies far outside training range, despite an overall tendency to underestimate large mass values and overestimate small mass values. In the springs domain, the PPN is able to predict objects with large spring charge relatively well but performs poorly on objects with low spring charge. This is likely due to the fact that objects with low spring charge tend to feel very little spring force overall, making the difference between charges of $32^{-1}$ and $16^{-1}$ much less noticeable than the difference between charges of $16$ and $32$.

\section{DISCUSSION}

We have presented the PPN, a model that is capable of discovering latent object properties in an entirely unsupervised manner from samples of object dynamics. Through our experiments, we showed not only that the representations of object properties learned by the PPN are sufficient to accurately simulate the dynamics of new systems under the same laws; but also that these learned representations can be readily transformed into relevant, human-interpretable properties such as mass and coefficient of restitution via principal component analysis.

The PPN demonstrates robustness by generalizing to novel scenarios with little loss in the accuracy of dynamical predictions or latent property inference. By using interaction networks as the basic building block of both our perception and prediction modules, we enabled our model to scale to arbitrary numbers of objects and interactions without architectural change. Our perception network architecture, in particular, is a simple but effective combination of relation and recurrent networks that may be useful in other time series inference tasks involving interacting objects. We also established the PPN's ability to infer latent properties outside the range of values seen during training, further boosting its potential in discovering the relevant latent properties of new systems.

Several extensions would further improve the applicability of our model to the general discovery of latent object properties. In particular, there are a few general classes of problems that which interaction network--based architectures haven't been able to solve: collision detection between rigid bodies of an arbitrary shape, dense fluid simulation, etc. Extending interaction networks to particle-based object representations is a promising future research direction~\cite{mrowca2018flexible}.

While the interaction network framework is generally extensible to arbitrary numbers of objects, the computational time required to process all objects scales quadratically with the number of objects due to the presence of interaction terms between all pairs of objects, making it impractical for very large systems. One way to improve the computational efficiency of both the perception and prediction modules is to only consider interactions from objects in the neighborhood of target objects (with the interpretation that most interactions are only strong on shorter length scales), similar to Chang et al. \cite{chang2016compositional}. A smaller, global interaction net could still be used to model longer range interactions.

The PPN provides a promising method for deriving the underlying properties governing the dynamics of systems, in addition to being a more general learnable physics engine capable of reasoning about potentially unknown object properties. The entirely unsupervised manner of its operation and its many generalization characteristics make the PPN suitable for application to a variety of systems, and it may even be able to discover relevant latent properties in domains that are yet to be well understood.
\section{ACKNOWLEDGMENTS}

We thank Michael Chang for his important insights and the anonymous reviewers for their useful suggestions. This work was supported by ONR MURI N00014-16-1-2007, the Center for Brain, Minds and Machines (NSF \#1231216), Facebook, and the Toyota Research Institute.

\bibliographystyle{unsrt}
\bibliography{references}

\end{document}